\documentclass{article}

\usepackage[nonatbib,final]{nips_2017}

\usepackage[round]{natbib}
\usepackage[utf8]{inputenc} 
\usepackage[T1]{fontenc}    
\usepackage[hyperfootnotes=false]{hyperref}       
\usepackage{url}            
\usepackage{booktabs}       
\usepackage{amsfonts}       
\usepackage{nicefrac}       
\usepackage{microtype}      

\usepackage{algorithm}
\usepackage{algpseudocode}
\usepackage{tabularx}
\usepackage{wrapfig}
\usepackage[toc,page]{appendix}

\usepackage{amsmath,amssymb,amstext,xparse}
\usepackage{textcmds}
\usepackage{float}
\usepackage{graphicx}
\makeatletter
\g@addto@macro\@floatboxreset\centering
\makeatother
\graphicspath{{./figures/}}
\DeclareMathOperator{\argmax}{argmax}

\renewcommand{\vec}[1]{\mathbf{#1}}

\title{Teacher-Student Curriculum Learning}

\author{
  Tambet Matiisen\thanks{Work done while interning at OpenAI.}\phantom{\footnotesize 1}\textsuperscript{,}\thanks{Correspondence to \texttt{tambet.matiisen@gmail.com}.} \\
  University of Tartu \\
  \And
  Avital Oliver\footnotemark[1]\phantom{\footnotesize 1}\textsuperscript{,}\thanks{Author currently at Google Brain} \\
  OpenAI \\
  \And
  Taco Cohen\footnotemark[1] \\
  University of Amsterdam \\
  \And
  John Schulman \\
  OpenAI \\
}

\begin{document}

\maketitle

\begin{abstract}

We propose Teacher-Student Curriculum Learning (TSCL), a framework for automatic curriculum learning, where the Student tries to learn a complex task and the Teacher automatically chooses subtasks from a given set for the Student to train on. We describe a family of Teacher algorithms that rely on the intuition that the Student should practice more those tasks on which it makes the fastest progress,\textit{ i.e.} where the slope of the learning curve is highest. In addition, the Teacher algorithms address the problem of forgetting by also choosing tasks where the Student's performance is getting worse.
We demonstrate that TSCL matches or surpasses
the results of carefully hand-crafted curricula in two tasks: addition of decimal numbers with LSTM and navigation in Minecraft. Using our automatically generated curriculum enabled to solve a Minecraft maze that could not be solved at all when training directly on solving the maze, and the learning was an order of magnitude faster than uniform sampling of subtasks.

\end{abstract}

\section{Introduction}

Deep reinforcement learning algorithms have been used to solve difficult tasks in video games \citep{Mnih2015}, locomotion \citep{Schulman2015, lillicrap2015continuous} and robotics \citep{Levine2015}. But tasks with sparse rewards like ``Robot, fetch me a beer'' remain challenging to solve with direct application of these algorithms. One reason is that the number of samples needed to solve a task with random exploration increases exponentially with the number of steps to get a reward \citep{langford2011efficient}. One approach to overcome this problem is to use curriculum learning \citep{Bengio2009,Zaremba2014,Graves2016,wu2017training}, where tasks are ordered by increasing difficulty and training only proceeds to harder tasks once easier ones are mastered. Curriculum learning helps when after mastering a simpler task the policy for a harder task is discoverable through random exploration.

To use curriculum learning, the researcher must:
\begin{itemize}
\item Be able to order subtasks by difficulty.
\item Decide on a \qq{mastery} threshold. This can be based on achieving certain score \citep{Zaremba2014,wu2017training}, which requires prior knowledge of acceptable performance of each task. Alternatively this can be based on a plateau of performance, which can be hard to detect given the noise in the learning curve.
\item Continuously mix in easier tasks while learning harder ones to avoid forgetting. Designing these mixtures effectively is challenging
\citep{Zaremba2014}.
\end{itemize} 

In this paper, we describe a new approach called Teacher-Student Curriculum Learning (TSCL). The Student is the model being trained. The Teacher monitors the Student's training progress and determines the tasks on which the Student should train at each training step, in order to maximize the Student's progression through the curriculum. The Student can be any machine learning model. The Teacher is itself learning about the Student as it's giving tasks, all as part of a single training session.

We describe several Teacher algorithms based on the notion of learning progress \citep{Oudeyer2007}. The main idea is that the Student should practice more the tasks on which it is making fastest progress i.e. the learning curve slope is highest. To counter forgetting, the Student should also practice tasks where the performance is getting worse
i.e. the learning curve slope is negative.

The main contributions of the paper are:
\begin{itemize}
    \item We formalize TSCL, a Teacher-Student framework for curriculum learning as partially observable Markov decision process (POMDP).
    \item We propose a family of algorithms based on the notion of learning progress. The algorithms also address the problem of forgetting previous tasks.
    \item We evaluate the algorithms on two supervised and reinforcement learning tasks: addition of decimal numbers with LSTM and navigation in Minecraft.
\end{itemize}

\section{Teacher-Student Setup}
\label{teacher-student-setup}

\begin{figure}[h!]
  \includegraphics[scale=0.5]{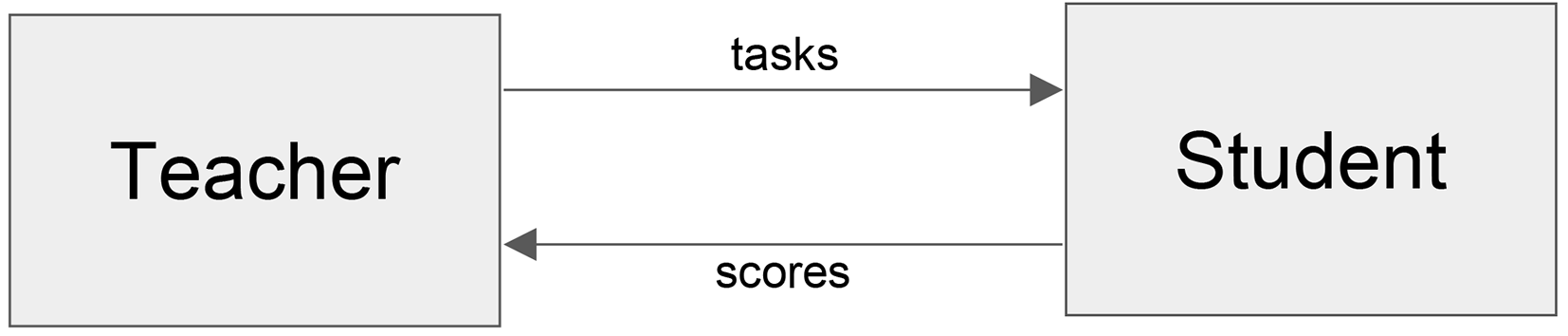}
\caption{The Teacher-Student setup}
\label{f1} 
\end{figure}

Figure \ref{f1} illustrates the Teacher-Student interaction.  At each timestep, the Teacher chooses tasks for the Student to practice on. The Student trains on those tasks and returns back a score. The Teacher's goal is for the Student to succeed on a final task with as few training steps as possible. Usually the task is parameterized by a categorical value representing one of $N$ subtasks, but one can imagine also multi-dimensional or continuous task parameterization. The score can be episode total reward in reinforcement learning or validation set accuracy in supervised learning.

We formalize the Teacher's goal of helping the Student to learn a final task as solving a partially observable Markov decision process (POMDP). We present two POMDP formulations: (1) \textit{Simple}, best suited for reinforcement learning; and (2) \textit{Batch}, best suited for supervised learning.

\subsection{Simple POMDP Formulation}
\label{ssec21}

The simple POMDP formulation exposes the score of the Student on a single task and is well-suited for reinforcement learning problems.
\begin{itemize}
\item  The state $s_t$ represents the entire state of the Student (\textit{i.e.} neural network parameters and optimizer state) and is not observable to the Teacher.

\item  The action $a_t$ corresponds to the parameters of the task chosen by Teacher. In following we only consider a discrete task parameterization. Taking an action means training Student on that task for certain number of iterations.

\item  The observation $o_t$ is the score $x^{(i)}_t$ of the task $i=a_t$ the Student trained on at timestep $t$, i.e. the episode total reward.
While in theory the Teacher could also observe other aspects of the Student state like network weights, for simplicity we choose to expose only the score.

\item  Reward $r_t$ is the change in score for the task the Student trained on at timestep $t$: $r_t=x^{(i)}_t-x^{(i)}_{t'_i}$, 
where $t'_i$ is the previous timestep when the same task was trained on.
\end{itemize}

\subsection{Batch POMDP Formulation}
\label{ssec22}

In supervised learning a training batch can include multiple tasks. Therefore action, observation, and reward apply to the whole training set and scores can be measured on a held-out validation set. This motivates the batch formulation of the POMDP:
\begin{itemize}
\item  The state $s_t$ represents training state of the Student.

\item  The action $a_t$ represents a probability distribution over $N$ tasks. Each training batch is sampled according to the distribution:
$
 a_t=(p^{(1)}_t,\ldots, p^{(N)}_t),
$
where $p^{(i)}_t$ is the probability of task $i$ at timestep $t$.

\item  The observation $o_t$ is the scores of all tasks after the training step:
$
o_t=(x^{(1)}_t,\ldots ,x^{(N)}_t)
$
In the simplest case the scores could be accuracies of the tasks in the training set. But in the case of minibatch training the model evolves during training and therefore additional evaluation pass is needed anyway to produce consistent results. Therefore we use a separate validation set that contains uniform mix of all tasks for this evaluation pass.

\item  The reward $r_t$ is the sum of changes in evaluation scores from the previous timestep:
$
r_t=\sum^N_{i=1}x^{(i)}_t-x^{(i)}_{t-1}
$.

\end{itemize}

This setup could also be used with reinforcement learning by performing training in batches of episodes. But because scoring one sample (one episode) in reinforcement learning is usually much more computationally expensive than in supervised learning, it makes sense to use simple POMDP formulation and make decision about the next task after each training step.

\subsection{Optimization Criteria}

For either of the POMDP formulations, maximizing the Teacher episode total reward is equivalent to maximizing the score of all tasks at the end of the episode:
$
\sum^T_{t=1}{}r_t=
\sum^N_{i=1}{}x^{(i)}_{T_i},
$
where $T_i$ is the last training step where task $i$ was being trained on\footnote{Due to telescoping summation cancelling out all $x_t^{(i)}$ terms but the $T_i$\textsuperscript{th}.}.

While an obvious choice for optimization criteria would have been the performance in the \textit{final task}, initially the Student might not have any success in the final task and this does not provide any meaningful feedback signal to the Teacher. Therefore we choose to maximize the sum of performances in \textit{all tasks}. The assumption here is that in curriculum learning the final task includes the elements of all previous tasks, therefore good performance in the intermediate tasks usually leads to good performance in the final task.

\section{Algorithms}

POMDPs are typically solved using reinforcement learning algorithms. But those require many training episodes, while we aim to train the Student in one Teacher episode. Therefore, we resort to simpler heuristics. The basic intuition is that the Student should practice those tasks more for which it is making most progress \citep{Oudeyer2007}, while also practicing tasks that are at risk of being forgotten.

\begin{figure}[h]
  \includegraphics[scale=0.65]{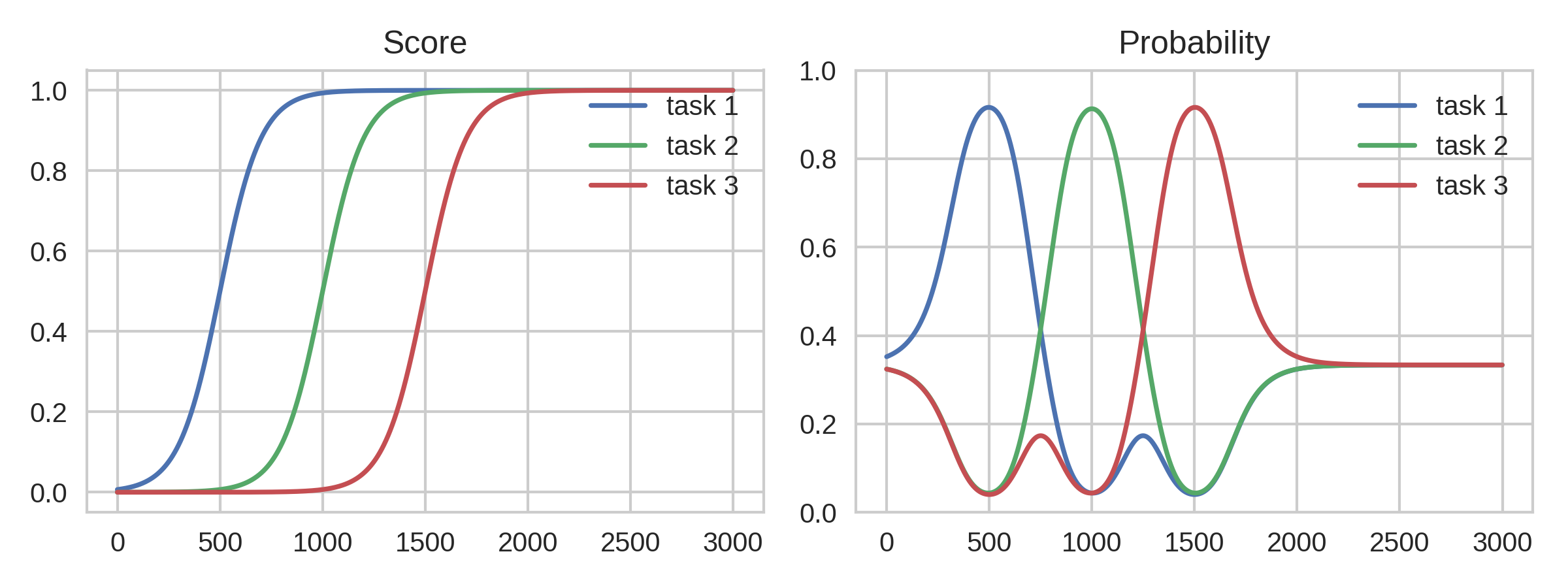}
\caption{Idealistic curriculum learning. Left: Scores of different tasks improve over time, the next task starts improving once the previous task has been mastered. Right: Probability of sampling a task depends on the slope of the learning curve.}
\label{f2}
\end{figure}

Figure \ref{f2} is a demonstration of the ideal training progress in a curriculum learning setting:
\begin{enumerate}
\item  At first, the Teacher has no knowledge so it samples from all tasks uniformly.

\item  When the Student starts making progress on task 1, the Teacher allocates more probability mass to this task.

\item  When the Student masters task 1, its learning curve flattens and the Teacher samples the task less often. At this point Student also starts making progress on task 2, so the Teacher samples more from task 2.

\item  This continues until the Student masters all tasks. As all task learning curves flatten in the end, the Teacher returns to uniform sampling of the tasks.
\end{enumerate}

The picture above is idealistic, since in practice some  unlearning often occurs, \textit{i.e.} when most of the probability mass is allocated to the task 2, performance on task 1 might get worse. To counter this the Student should also practice all learned tasks, especially those where unlearning occurs. For this reason we sample tasks according to the \textit{absolute value} of the slope of the learning curve instead. If the change in scores is negative, this must mean that unlearning occurred and this task should be practiced more.

This description alone does not prescribe an algorithm. We need to propose a method of estimating learning progress from noisy task scores, and a way to balance exploration and exploitation. We take inspiration from algorithms for the non-stationary multi-armed bandit problem \citep{Sutton1998} and adapt them to TSCL. For brevity we only give intuition for the simple formulation algorithms here, the formal descriptions can be found in appendices \ref{appendix:simple_algs} and \ref{appendix:batch_algs}.

\subsection{Online algorithm}

The Online algorithm is inspired by the basic non-stationary bandit algorithm
\citep{Sutton1998}. It uses exponentially weighted moving average to track the expected return $Q$ from different tasks:
\[
Q_{t+1}(a_t)=\alpha r_t+(1- \alpha)Q_t(a_t),
\]
where $\alpha$ is learning rate. The next task can be chosen by $\epsilon$-greedy exploration: sample a random task with probability $\epsilon$, or $\argmax Q_t(a)$ otherwise.

Alternatively the next task can be chosen using Boltzmann distribution:
\[
p(a)=\frac{e^{Q_t(a)/\tau}}{\sum^N_{i=1}e^{Q_t(i)/\tau}},
\]
where $\tau$ is the temperature of Boltzmann distribution. For details, see Algorithm \ref{online_simple} in Appendix \ref{appendix:simple_algs}.

\subsection{Naive algorithm}

To estimate the learning progress more reliably one should practice the task several times. The Naive algorithm trains each task $K$ times, observes the resulting scores and estimates the slope of the learning curve using linear regression. The regression coefficient is used as the reward in the above non-stationary bandit algorithm. For details, see Algorithm \ref{naive_simple} in Appendix \ref{appendix:simple_algs}.

\subsection{Window algorithm}

Repeating the task a fixed number of times is expensive, when clearly no progress is made. The Window algorithm keeps FIFO buffer of last $K$ scores, and timesteps when these scores were recorded. Linear regression is performed to estimate the slope of the learning curve for each task, with the timesteps as the input variables. The regression coefficient is used as the reward in the above non-stationary bandit algorithm. For details, see Algorithm \ref{window_simple} in Appendix \ref{appendix:simple_algs}.
  
\subsection{Sampling algorithm}

The previous algorithms require tuning of hyperparameters to balance exploration. To get rid of exploration hyperparameters, we take inspiration from Thompson sampling. The Sampling algorithm keeps a buffer of last $K$ rewards for each task. To choose the next task, a recent reward is sampled from each task's $K$-last-rewards buffer.  Then whichever task yielded the highest sampled reward is chosen. This makes exploration a natural part of the algorithm: tasks that have recently had high rewards are sampled more often. For details, see Algorithm \ref{sampling_simple} in Appendix \ref{appendix:simple_algs}.

\section{Experiments}

\subsection{Decimal Number Addition}

Addition of decimal numbers with LSTM is a well known task that requires a curriculum to learn in reasonable time \citep{Zaremba2014}. It is implemented as sequence-to-sequence model \citep{sutskever2014sequence}, where the input to the network is two decimal-coded numbers separated by a 'plus' sign, and output of the network is the sum of those numbers, also in decimal coding. The curriculum is based on the number of digits in the input numbers -- it is easier to learn addition of short numbers and then move on to longer numbers.

Number addition is a supervised learning problem and therefore can be trained more efficiently by including several curriculum tasks in the mini-batch. Therefore we adopt batch training scheme as outlined in \ref{ssec22}. The score we use is the accuracy of each task calculated on validation set. The results shown below are means and standard deviations of 3 runs with different random seeds. Full experiment details can be found in appendix \ref{appendix:addition}.

\subsubsection{Addition with 1-dimensional Curriculum}

\begin{figure}[h]
\begin{minipage}[b]{65mm}
  \includegraphics[scale=0.42]{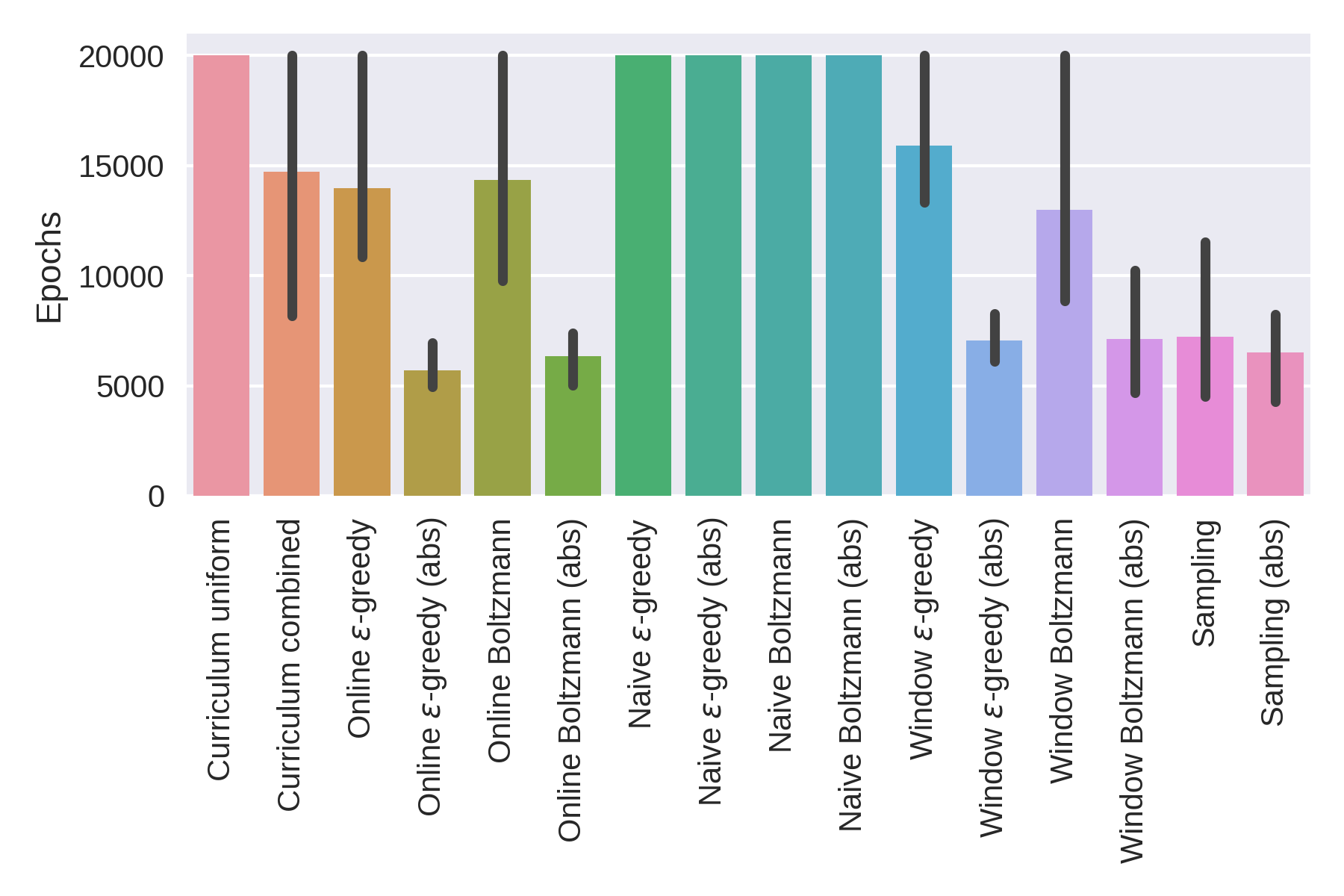}
\caption{Results for 9-digit 1D addition, lower is better. Variants using the absolute value of the expected reward surpass the best manual curriculum ("combined").}
\label{f5}
\end{minipage}
\hfill
\begin{minipage}[b]{65mm}
  \includegraphics[scale=0.39]{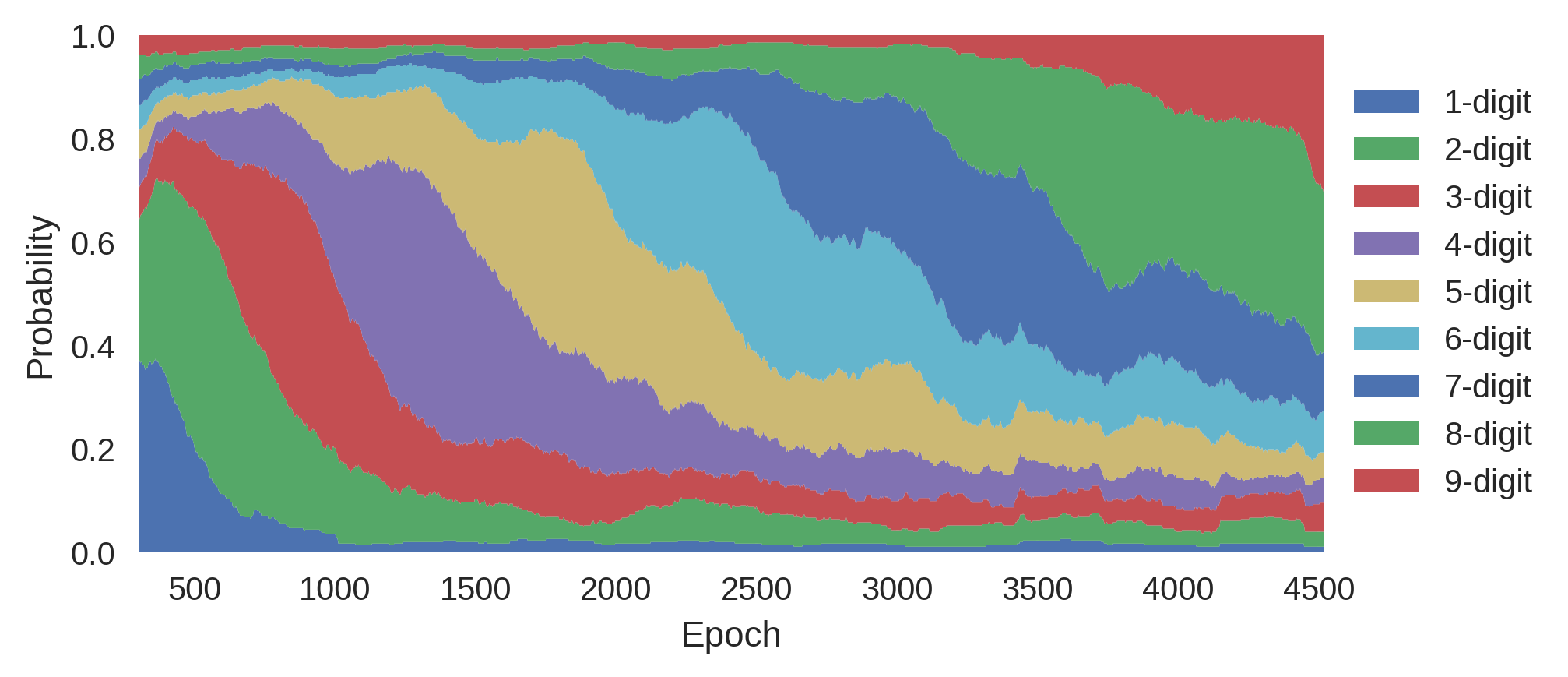}
\caption{Progression of the task distribution over time for 9-digit 1D addition (Sampling). The algorithm progresses from simpler tasks to more complicated. Harder tasks take longer to learn and the algorithm keeps training on easier tasks to counter unlearning.}
\label{f6}
\end{minipage}
\end{figure}

We started with a similar setup to \citep{Zaremba2014}, where the curriculum task determines the maximum number of digits in both added numbers. The results are shown on Figure \ref{f5}. Our algorithms outperformed uniform sampling and the best manual curriculum ("combined") for 9-digit addition from \citep{Zaremba2014}. An example of the task distribution during training session is given on figure \ref{f6}.

\subsubsection{Addition with 2-dimensional Curriculum}

We also experimented with a curriculum where the ordering of tasks is not obvious. We used the same decimal addition task, but in this case the length of each number is chosen separately, making the task-space 2-dimensional. Each training batch is modelled as a probability distribution over the length of both numbers $P(l_1, l_2)$. We also tried making this distribution independent such that $P(l_1, l_2) = P(l_1) P(l_2)$, but that did not work as well.

\begin{figure}[h]
\begin{minipage}[b]{65mm}
  \includegraphics[scale=0.42]{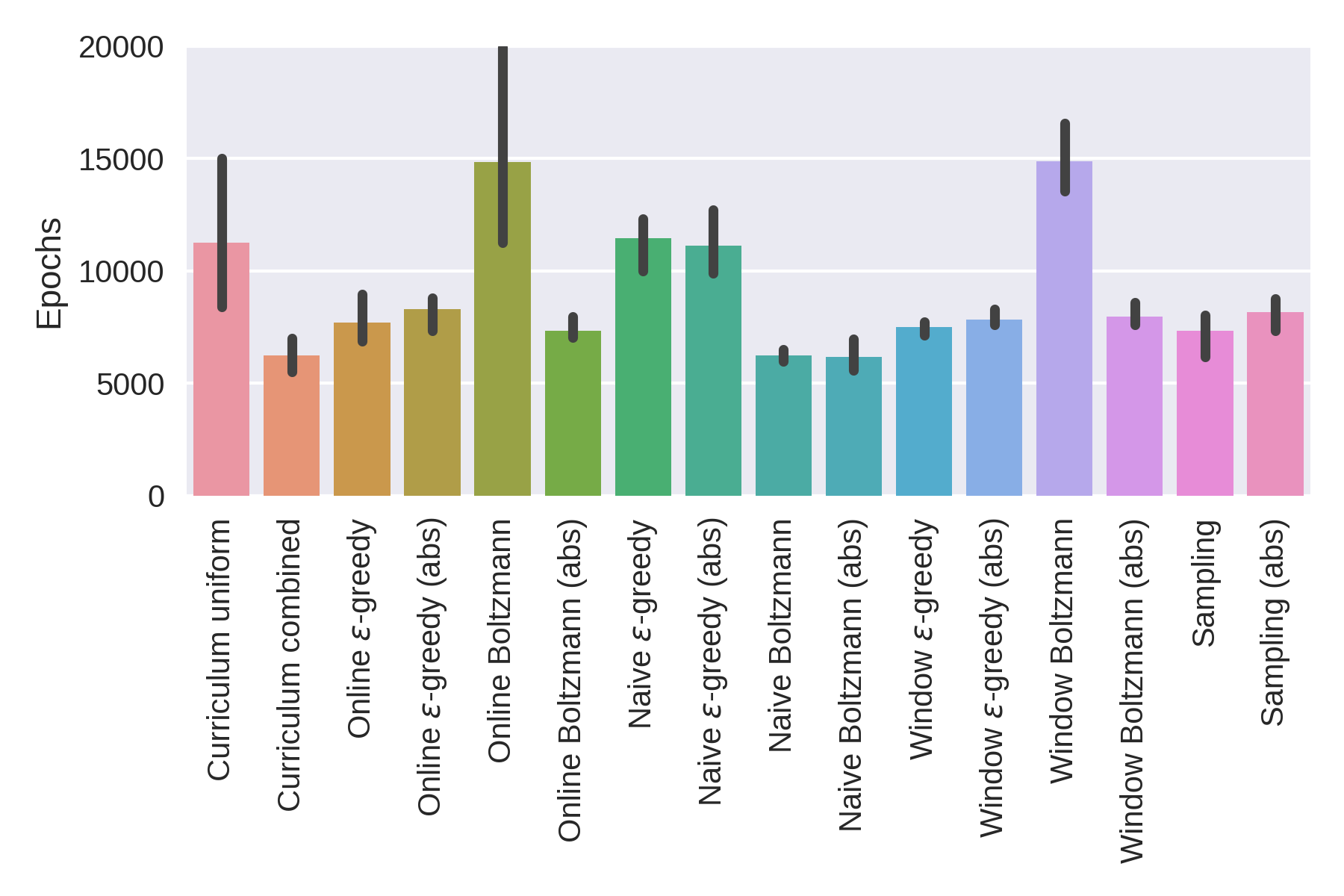}
\caption{Results for 9-digit 2D addition, lower is better. The task seems easier, manual curriculum is hard to beat and uniform sampling is competitive.}
\label{f7}
\end{minipage}
\hfill
\begin{minipage}[b]{65mm}
  \includegraphics[scale=0.28]{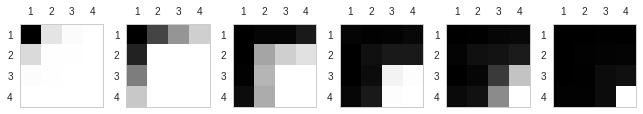}
  \includegraphics[scale=0.28]{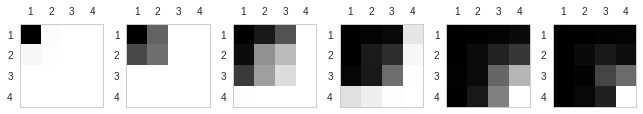}
\caption{Accuracy progress for 4-digit 2D addition. Top: TSCL. Bottom: the best manual curriculum. Our algorithm takes distinctively different approach by training on shorter numbers first. 9-digit videos can be found \url{https://youtu.be/y_QIcQ6spWk} and \url{https://youtu.be/fB2kx-esjgw}.}
\label{f8}
\end{minipage}
\end{figure}

There is no equivalent experiment in \citep{Zaremba2014}, so we created a manual curriculum inspired by their best 1D curriculum. In particular we increase difficulty by increasing the maximum length of both two numbers, which reduces the problem to a 1D curriculum. Figure \ref{f7} shows the results for 9-digit 2D addition. Figure \ref{f8} illustrates the different approaches taken by manual and automated curriculum.

\subsubsection{Observations}

\begin{itemize}
\item Using absolute value of $Q$ boosts the performance of almost all the algorithms, which means it is efficient in countering forgetting.

\item There is no universal best algorithm. For 1D the Window algorithm and for 2D the Naive algorithm performed the best. Sampling is competitive in both and has least hyperparameters.

\item Whether $\epsilon$-greedy or Boltzmann exploration works better depends on the algorithm.

\item Uniform sampling is surprisingly efficient, especially in 2D case.

\item The 2D task is solved faster and the manual curriculum is hard to beat in 2D.
\end{itemize}

\subsection{Minecraft}

Minecraft is a popular 3D video game where players can explore, craft tools and build arbitrary structures, making it a potentially rich environment for AI research. We used the Malmo platform \citep{johnson2016malmo} with OpenAI Gym wrapper\footnote{\url{ https://github.com/tambetm/gym-minecraft}}
to interact with Minecraft in our reinforcement learning experiments. In particular we used \textit{ClassroomDecorator}
from Malmo to generate random mazes for the agent to solve. The mazes contain sequences of rooms separated by the following obstacles:
\begin{itemize}
\item  \textbf{Wall} -- the agent has to locate a doorway in the wall.

\item  \textbf{Lava} -- the agent has to cross a bridge over lava.



\end{itemize}

We only implemented the Window algorithm for the Minecraft task, because other algorithms rely on score change, which is not straightforward to calculate for parallel training scheme. As baseline we use uniform sampling, training only on the last task, and a manually tuned curriculum. Full experimental details can be found in appendix \ref{appendix:minecraft}.

\begin{wrapfigure}[3]{r}{0.5\textwidth}
  \begin{center}
    \includegraphics[width=0.4\textwidth]{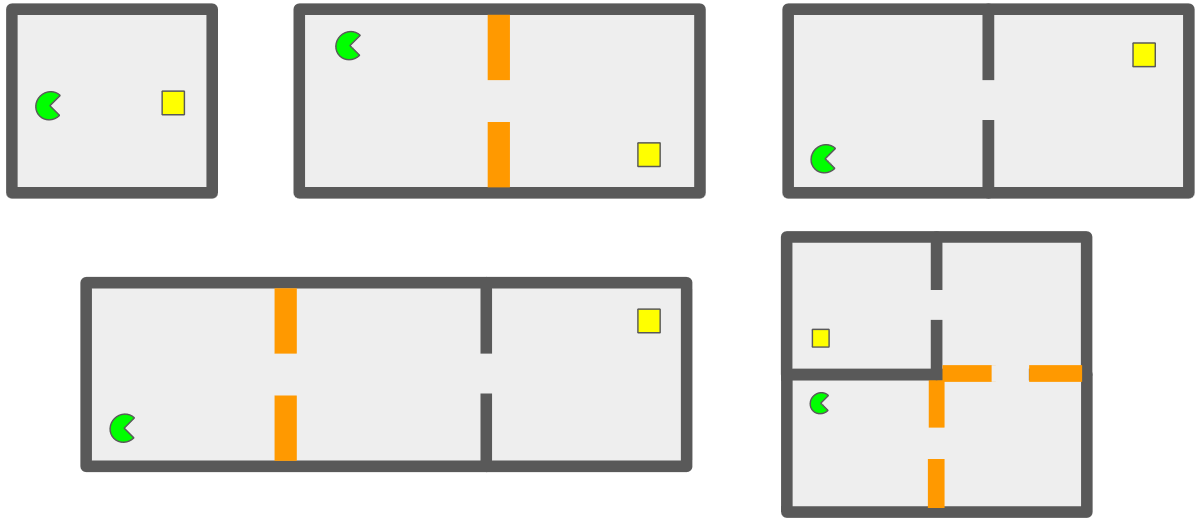}
  \end{center}
  \caption{5-step curriculum.}
  \label{5step}
\end{wrapfigure}

\subsubsection{5-step Curriculum}

We created a simple curriculum with 5 steps:

\begin{enumerate}
\item 
A single room with a target.

\item  
Two rooms separated by lava.

\item  
Two rooms separated by wall.

\item  
Three rooms separated by lava and wall, in random order.

\item  
Four rooms separated by lava and walls, in random order.
\end{enumerate}

Refer to Figure \ref{5step} for the room layout. The starting position of the agent and the location of the target were randomized for each episode. Manual curriculum trained first task for $200\,000$ steps, second, third and fourth task for $400\,000$ steps, and fifth task for $600\,000$ steps.

\begin{figure}[h]
  \includegraphics[scale=0.5]{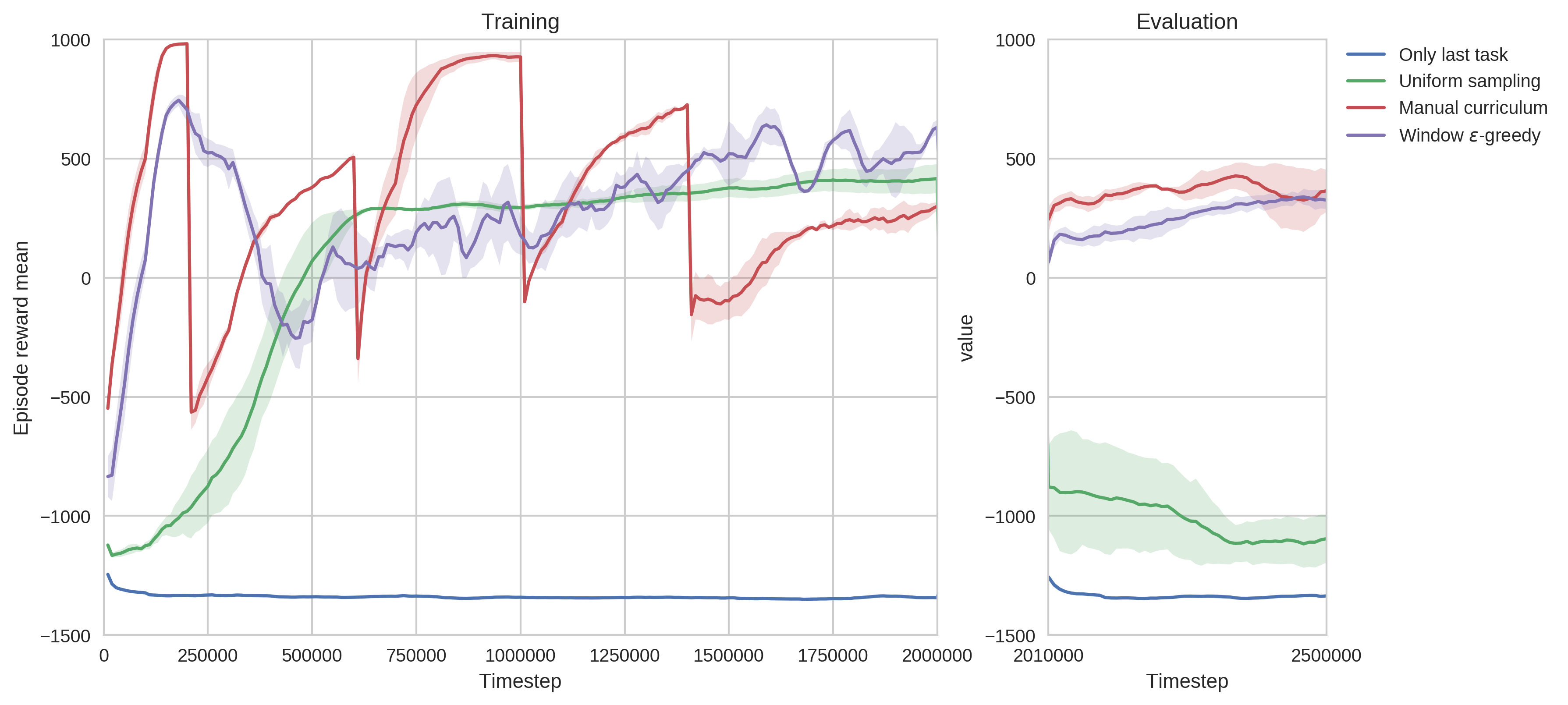}
\caption{Minecraft 5-step curriculum results, Y-axis shows mean episode reward per $10\,000$ timesteps for the current task. Left: training performance, notice the manual curriculum task switches after $200\,000$, $600\,000$, $1\,000\,000$ and $1\,400\,000$ steps. For automatic curriculum the training score has no clear interpretation. Right: evaluation training on the last task. When training only on the last task the agent did not make any progress at all. When training on a uniform mix of the tasks the progress was slow. Manual curriculum allowed the agent to learn the last task to an acceptable level. TSCL is comparable to the manual curriculum in performance.}
\label{f11}
\end{figure}

Figure \ref{f11} shows learning curves for Minecraft 5-step curriculum. The mean curve and standard deviation are based on 3 runs with different random seeds.

Video of the trained agent can be found here: \url{https://youtu.be/cada0d_aDIc}. The learned policy is robust to the number of rooms, given that obstacles are of the same type. The code is available at \url{https://github.com/tambetm/TSCL}.

\section{Related Work}

Work by \citep{Bengio2009} sparked general interest in curriculum learning. More recent results include learning to execute short programs \citep{Zaremba2014}, finding shortest paths in graphs \citep{Graves2016} and learning to play first-person shooter \citep{wu2017training}. All those works rely on manually designed curricula and do not attempt to produce it automatically.

The idea of using learning progress as the reward could be traced back to \citep{schmidhuber1991curious}. It has been successfully applied in the context of developmental robotics to learn object manipulation \citep{oudeyer2007intrinsic,baranes2013active} and also in actual classroom settings to teach primary school students \citep{clement2015multi}. Using learning progress as the reward can be linked to the concept of intrinsic motivation \citep{Oudeyer2007,Schmidhuber2010}.

Several algorithms for adversarial bandits were analyzed in \citep{Auer2002}. While many of those algorithms have formal worst-case guarantees, in our experiments they did not perform well. The problem is that they come with no assumptions. In curriculum learning we can assume that rewards change smoothly over time.

More recently \citep{sukhbaatar2017intrinsic} proposed a method to generate incremental goals and therefore curricula automatically. The setup consists of two agents, Alice and Bob, where Alice is generating trajectories and Bob is trying to either repeat or reverse them. Similar work by \citep{held2017automatic} uses generative adversarial network to generate goal states for an agent. Compared to TSCL, they are able to generate new subtasks on the go, but this mainly aids in exploration and is not guaranteed to help in learning the final task. \citep{sharma2017online} apply similar setup as ours to multi-task learning. In their work they practice more tasks that are underperforming compared to preset baseline, as opposed to our approach of using learning progress. \citep{jain2017faster} estimate transfer between subtasks and target task, and create curriculum based on that.

The most similar work to ours was done concurrently in \citep{Graves2017}. While the problem statement is strikingly similar, our approaches differ. They apply the automatic curriculum learning only to supervised sequence learning tasks, while we consider also reinforcement learning tasks. They use the EXP3.S algorithm for adversarial bandits, while we propose alternative algorithms inspired by non-stationary bandits. They consider other learning progress metrics based on complexity gain while we focus only on prediction gain (which performed overall best in their experiments). Moreover, their work only uses uniform sampling of tasks as a baseline, whereas ours compares the best known manual curriculum for the given tasks. In summary they arrive to very similar conclusions to ours.

Decimal addition has also been explored in \citep{kalchbrenner2015grid, reed2015neural, kaiser2015neural}, sometimes improving results over original work in \citep{Zaremba2014}. Our goal was not to improve the addition results, but to evaluate different curriculum approaches, therefore there is no direct comparison.

Minecraft is a relatively recent addition to reinforcement learning environments. Work by \citep{oh2016control} evaluates memory-based architectures for Minecraft. They use cognition-inspired tasks in visual grid-world. Our tasks differ in that they do not need explicit memory, and the movement is continuous, not grid-world. Another work by \citep{tessler2016deep} uses tasks similar to ours but they take different approach: they learn a Deep Skill Module for each subtask, freeze weights of those modules and train hierarchical deep reinforcement learning network to pick either single actions or subtask policies. In contrast our approach uses simple policy network and relies on the TSCL to learn (and not forget) the subtasks.

While exploration bonuses \citep{bellemare2016unifying,houthooft2016vime,stadie2015incentivizing} solve the same problem of sparse rewards, they apply to Student algorithms, while we were considering different Teacher approaches. For this reason we leave the comparison with exploration bonuses to future work.


\section{Conclusion}
We presented a framework for automatic curriculum learning that can be used for supervised and reinforcement learning tasks. We proposed a family of algorithms within that framework based on the concept of learning progress. While many of the algorithms performed equally well, it was crucial to rely on the absolute value of the slope of the learning curve when choosing the tasks. This guarantees the re-training on tasks which the network is starting to forget. In our LSTM decimal addition experiments, the Sampling algorithm outperformed the best manually designed curriculum as well as the uniform sampling. On the challenging 5-task Minecraft navigation problem, our Window algorithm matched the performance of a carefully designed manual curriculum, and significantly outperformed uniform sampling. For problems where curriculum learning is necessary, TSCL can avoid the tedium of ordering the difficulty of subtasks and hand-designing the curriculum.

\section{Future Work}
In this work we only considered discrete task parameterizations. In the future it would be interesting to apply the idea to continuous task parameterizations. Another promising idea to explore is the usage of automatic curriculum learning in contexts where the subtasks have not been pre-defined. For example, subtasks can be sampled from a generative model, or taken from different initial states in the same environment.

\section{Acknowledgements}
We thank Microsoft for their excellent Malmö environment for Minecraft, Josh Tobin and Pieter Abbeel for suggestions and comments, Vicky Cheung, Jonas Schneider, Ben Mann and Art Chaidarun for always being helpful with OpenAI infrastructure. Also Raul Vicente, Ardi Tampuu and Ilya Kuzovkin from University of Tartu for comments and discussion.
\nocite{*}

\bibliographystyle{plainnat}
\bibliography{main}

\begin{thebibliography}{37}
\providecommand{\natexlab}[1]{#1}
\providecommand{\url}[1]{\texttt{#1}}
\expandafter\ifx\csname urlstyle\endcsname\relax
  \providecommand{\doi}[1]{doi: #1}\else
  \providecommand{\doi}{doi: \begingroup \urlstyle{rm}\Url}\fi

\bibitem[Auer et~al.(2002)Auer, Cesa-Bianchi, Freund, and Schapire.]{Auer2002}
Peter Auer, Nicolò‚ Cesa-Bianchi, Yoav Freund, and Robert~E. Schapire.
\newblock The non-stochastic multi-armed bandit problem.
\newblock \emph{SIAM Journal on Computing}, 32\penalty0 (1):\penalty0 48--77,
  2002.

\bibitem[Babaeizadeh et~al.(2016)Babaeizadeh, Frosio, Tyree, Clemons, and
  Kautz]{babaeizadeh2016reinforcement}
Mohammad Babaeizadeh, Iuri Frosio, Stephen Tyree, Jason Clemons, and Jan Kautz.
\newblock Reinforcement learning through asynchronous advantage actor-critic on
  a gpu.
\newblock 2016.

\bibitem[Baranes and Oudeyer(2013)]{baranes2013active}
Adrien Baranes and Pierre-Yves Oudeyer.
\newblock Active learning of inverse models with intrinsically motivated goal
  exploration in robots.
\newblock \emph{Robotics and Autonomous Systems}, 61\penalty0 (1):\penalty0
  49--73, 2013.

\bibitem[Bellemare et~al.(2016)Bellemare, Srinivasan, Ostrovski, Schaul,
  Saxton, and Munos]{bellemare2016unifying}
Marc Bellemare, Sriram Srinivasan, Georg Ostrovski, Tom Schaul, David Saxton,
  and Remi Munos.
\newblock Unifying count-based exploration and intrinsic motivation.
\newblock In \emph{Advances in Neural Information Processing Systems}, pages
  1471--1479, 2016.

\bibitem[Bengio et~al.(2009)Bengio, Louradour, Collobert, and
  Weston]{Bengio2009}
Yoshua Bengio, Jérôme Louradour, Ronan Collobert, and Jason Weston.
\newblock Curriculum learning.
\newblock In \emph{Proceedings of the 26th Annual International Conference on
  Machine Learning -- ICML '09}, 2009.
\newblock \doi{10.1145/1553374.1553380.}

\bibitem[Chollet et~al.(2015)]{chollet2015keras}
Fran\c{c}ois Chollet et~al.
\newblock Keras.
\newblock \url{https://github.com/fchollet/keras}, 2015.

\bibitem[Clement et~al.(2015)Clement, Roy, Oudeyer, and
  Lopes]{clement2015multi}
Benjamin Clement, Didier Roy, Pierre-Yves Oudeyer, and Manuel Lopes.
\newblock Multi-armed bandits for intelligent tutoring systems.
\newblock \emph{Journal of Educational Data Mining (JEDM)}, 7\penalty0 (2),
  2015.

\bibitem[Graves et~al.(2016)Graves, Wayne, Reynolds, Harley, Danihelka,
  Grabska-Barwińska, and Sergio Gómez~Colmenarejo]{Graves2016}
Alex Graves, Greg Wayne, Malcolm Reynolds, Tim Harley, Ivo Danihelka, Agnieszka
  Grabska-Barwińska, and et~al. Sergio Gómez~Colmenarejo.
\newblock Hybrid computing using a neural network with dynamic external memory.
\newblock \emph{Nature}, 538\penalty0 (7626):\penalty0 71--76, 2016.

\bibitem[Graves et~al.(2017)Graves, Bellemare, Menick, Munos, and
  Kavukcuoglu]{Graves2017}
Alex Graves, Marc~G. Bellemare, Jacob Menick, Remi Munos, and Koray
  Kavukcuoglu.
\newblock Automated curriculum learning for neural networks, 2017.
\newblock http://arxiv.org/abs/1704.03003.

\bibitem[Held et~al.(2017)Held, Geng, Florensa, and Abbeel]{held2017automatic}
David Held, Xinyang Geng, Carlos Florensa, and Pieter Abbeel.
\newblock Automatic goal generation for reinforcement learning agents.
\newblock \emph{arXiv preprint arXiv:1705.06366}, 2017.

\bibitem[Houthooft et~al.(2016)Houthooft, Chen, Duan, Schulman, De~Turck, and
  Abbeel]{houthooft2016vime}
Rein Houthooft, Xi~Chen, Yan Duan, John Schulman, Filip De~Turck, and Pieter
  Abbeel.
\newblock Vime: Variational information maximizing exploration.
\newblock In \emph{Advances in Neural Information Processing Systems}, pages
  1109--1117, 2016.

\bibitem[Jain and Tulabandhula(2017)]{jain2017faster}
Vikas Jain and Theja Tulabandhula.
\newblock Faster reinforcement learning using active simulators.
\newblock \emph{arXiv preprint arXiv:1703.07853}, 2017.

\bibitem[Johnson et~al.(2016)Johnson, Hofmann, Hutton, and
  Bignell]{johnson2016malmo}
Matthew Johnson, Katja Hofmann, Tim Hutton, and David Bignell.
\newblock The malmo platform for artificial intelligence experimentation.
\newblock In \emph{International joint conference on artificial intelligence
  (IJCAI)}, page 4246, 2016.

\bibitem[Kaiser and Sutskever(2015)]{kaiser2015neural}
{\L}ukasz Kaiser and Ilya Sutskever.
\newblock Neural gpus learn algorithms.
\newblock \emph{arXiv preprint arXiv:1511.08228}, 2015.

\bibitem[Kalchbrenner et~al.(2015)Kalchbrenner, Danihelka, and
  Graves]{kalchbrenner2015grid}
Nal Kalchbrenner, Ivo Danihelka, and Alex Graves.
\newblock Grid long short-term memory.
\newblock \emph{arXiv preprint arXiv:1507.01526}, 2015.

\bibitem[Kingma and Ba(2014)]{kingma2014adam}
Diederik Kingma and Jimmy Ba.
\newblock Adam: A method for stochastic optimization.
\newblock \emph{arXiv preprint arXiv:1412.6980}, 2014.

\bibitem[Langford(2011)]{langford2011efficient}
John Langford.
\newblock Efficient exploration in reinforcement learning.
\newblock In \emph{Encyclopedia of Machine Learning}, pages 309--311. Springer,
  2011.

\bibitem[Levine et~al.(2015)Levine, Finn, Darrell, and Abbeel]{Levine2015}
Sergey Levine, Chelsea Finn, Trevor Darrell, and Pieter Abbeel.
\newblock End-to-end training of deep visuomotor policies, 2015.
\newblock http://arxiv.org/abs/1504.00702.

\bibitem[Lillicrap et~al.(2015)Lillicrap, Hunt, Pritzel, Heess, Erez, Tassa,
  Silver, and Wierstra]{lillicrap2015continuous}
Timothy~P Lillicrap, Jonathan~J Hunt, Alexander Pritzel, Nicolas Heess, Tom
  Erez, Yuval Tassa, David Silver, and Daan Wierstra.
\newblock Continuous control with deep reinforcement learning.
\newblock \emph{arXiv preprint arXiv:1509.02971}, 2015.

\bibitem[Mirowski et~al.(2016)Mirowski, Pascanu, Viola, Soyer, Ballard, Banino,
  Denil, Goroshin, Sifre, Kavukcuoglu, et~al.]{mirowski2016learning}
Piotr Mirowski, Razvan Pascanu, Fabio Viola, Hubert Soyer, Andy Ballard, Andrea
  Banino, Misha Denil, Ross Goroshin, Laurent Sifre, Koray Kavukcuoglu, et~al.
\newblock Learning to navigate in complex environments.
\newblock \emph{arXiv preprint arXiv:1611.03673}, 2016.

\bibitem[Mnih et~al.(2015)Mnih, Kavukcuoglu, Silver, Rusu, Veness, Bellemare,
  and Alex~Graves]{Mnih2015}
Volodymyr Mnih, Koray Kavukcuoglu, David Silver, Andrei~A. Rusu, Joel Veness,
  Marc~G. Bellemare, and et~al. Alex~Graves.
\newblock Human-level control through deep reinforcement learning.
\newblock \emph{Nature}, 518\penalty0 (7540):\penalty0 529--33, 2015.

\bibitem[Oh et~al.(2016)Oh, Chockalingam, Singh, and Lee]{oh2016control}
Junhyuk Oh, Valliappa Chockalingam, Satinder Singh, and Honglak Lee.
\newblock Control of memory, active perception, and action in minecraft.
\newblock \emph{arXiv preprint arXiv:1605.09128}, 2016.

\bibitem[Oudeyer and Kaplan(2007)]{Oudeyer2007}
Pierre-Yves Oudeyer and Frederic Kaplan.
\newblock What is intrinsic motivation? a typology of computational approaches.
\newblock \emph{Frontiers in Neurorobotics}, 1\penalty0 (November: 6), 2007.

\bibitem[Oudeyer et~al.(2007)Oudeyer, Kaplan, and Hafner]{oudeyer2007intrinsic}
Pierre-Yves Oudeyer, Frdric Kaplan, and Verena~V Hafner.
\newblock Intrinsic motivation systems for autonomous mental development.
\newblock \emph{IEEE transactions on evolutionary computation}, 11\penalty0
  (2):\penalty0 265--286, 2007.

\bibitem[Reed and De~Freitas(2015)]{reed2015neural}
Scott Reed and Nando De~Freitas.
\newblock Neural programmer-interpreters.
\newblock \emph{arXiv preprint arXiv:1511.06279}, 2015.

\bibitem[Schmidhuber(1991)]{schmidhuber1991curious}
J{\"u}rgen Schmidhuber.
\newblock Curious model-building control systems.
\newblock In \emph{Neural Networks, 1991. 1991 IEEE International Joint
  Conference on}, pages 1458--1463. IEEE, 1991.

\bibitem[Schmidhuber(2010)]{Schmidhuber2010}
Jürgen Schmidhuber.
\newblock Formal theory of creativity, fun, and intrinsic motivation
  (1990--2010).
\newblock \emph{IEEE Transactions on Autonomous Mental Development}, 2\penalty0
  (3):\penalty0 230--47, 2010.

\bibitem[Schulman et~al.(2015)Schulman, Levine, Abbeel, Jordan, and
  Moritz]{Schulman2015}
John Schulman, Sergey Levine, Pieter Abbeel, Michael~I Jordan, and Philipp
  Moritz.
\newblock Trust region policy optimization.
\newblock In \emph{ICML}, pages 1889--1897, 2015.

\bibitem[Schulman et~al.(2017)Schulman, Wolski, Dhariwal, Radford, and
  Klimov]{schulman2017proximal}
John Schulman, Filip Wolski, Prafulla Dhariwal, Alec Radford, and Oleg Klimov.
\newblock Proximal policy optimization algorithms.
\newblock \emph{arXiv preprint arXiv:1707.06347}, 2017.

\bibitem[Sharma and Ravindran(2017)]{sharma2017online}
Sahil Sharma and Balaraman Ravindran.
\newblock Online multi-task learning using biased sampling.
\newblock \emph{arXiv preprint arXiv:1702.06053}, 2017.

\bibitem[Stadie et~al.(2015)Stadie, Levine, and
  Abbeel]{stadie2015incentivizing}
Bradly~C Stadie, Sergey Levine, and Pieter Abbeel.
\newblock Incentivizing exploration in reinforcement learning with deep
  predictive models.
\newblock \emph{arXiv preprint arXiv:1507.00814}, 2015.

\bibitem[Sukhbaatar et~al.(2017)Sukhbaatar, Kostrikov, Szlam, and
  Fergus]{sukhbaatar2017intrinsic}
Sainbayar Sukhbaatar, Ilya Kostrikov, Arthur Szlam, and Rob Fergus.
\newblock Intrinsic motivation and automatic curricula via asymmetric
  self-play.
\newblock \emph{arXiv preprint arXiv:1703.05407}, 2017.

\bibitem[Sutskever et~al.(2014)Sutskever, Vinyals, and
  Le]{sutskever2014sequence}
Ilya Sutskever, Oriol Vinyals, and Quoc~V Le.
\newblock Sequence to sequence learning with neural networks.
\newblock In \emph{Advances in neural information processing systems}, pages
  3104--3112, 2014.

\bibitem[Sutton and Barto(1998)]{Sutton1998}
Richard~S. Sutton and Andrew~G. Barto.
\newblock Reinforcement learning: An introduction.
\newblock \emph{IEEE Transactions on Neural Networks / a Publication of the
  IEEE Neural Networks Council}, 9\penalty0 (5):\penalty0 1054--1054, 1998.

\bibitem[Tessler et~al.(2016)Tessler, Givony, Zahavy, Mankowitz, and
  Mannor]{tessler2016deep}
Chen Tessler, Shahar Givony, Tom Zahavy, Daniel~J Mankowitz, and Shie Mannor.
\newblock A deep hierarchical approach to lifelong learning in minecraft.
\newblock \emph{arXiv preprint arXiv:1604.07255}, 2016.

\bibitem[Wu and Tian(2017)]{wu2017training}
Yuxin Wu and Yuandong Tian.
\newblock Training agent for first-person shooter game with actor-critic
  curriculum learning.
\newblock In \emph{Submitted to Int’l Conference on Learning
  Representations}, 2017.

\bibitem[Zaremba and Sutskever(2014)]{Zaremba2014}
Wojciech Zaremba and Ilya Sutskever.
\newblock Learning to execute., 2014.
\newblock http://arxiv.org/abs/1410.4615.

\end{thebibliography}

\clearpage
\begin{appendices}

\section{Simple versions of the algorithms}
\label{appendix:simple_algs}

\begin{algorithm}
\caption{Online algorithm}\label{online_simple}
\begin{algorithmic}
\State Initialize \textsc{Student} learning algorithm
\State Initialize expected return $Q(a)=0$ for all $N$ tasks
\For{t=1,\ldots,T}
\State Choose task $a_t$ based on $|Q|$ using $\epsilon$-greedy or Boltzmann policy
\State Train \textsc{Student} using task $a_t$ and observe reward $r_t = x_t^{(a_t)} - x_{t'}^{(a_t)}$
\State Update expected return $Q(a_t) = \alpha r_t + (1 - \alpha) Q(a_t)$
\EndFor
\end{algorithmic}
\end{algorithm}

\begin{algorithm}
\caption{Naive algorithm}\label{naive_simple}
\begin{algorithmic}
\State Initialize \textsc{Student} learning algorithm
\State Initialize expected return $Q(a)=0$ for all $N$ tasks
\For{t=1,...,T}
\State Choose task $a_t$ based on $|Q|$ using $\epsilon$-greedy or Boltzmann policy
\State Reset $D=\emptyset$
\For{k=1,...,K}
\State Train \textsc{Student} using task $a_t$ and observe score $o_t = x_t^{(a_t)}$
\State Store score $o_t$ in list $D$
\EndFor
\State Apply linear regression to $D$ and extract the coefficient as $r_t$
\State Update expected return $Q(a_t) = \alpha r_t + (1 - \alpha) Q(a_t)$
\EndFor
\end{algorithmic}
\end{algorithm}

\begin{algorithm}
\caption{Window algorithm}\label{window_simple}
\begin{algorithmic}
\State Initialize \textsc{Student} learning algorithm
\State Initialize FIFO buffers $D(a)$ and $E(a)$ with length $K$ for all $N$ tasks
\State Initialize expected return $Q(a)=0$ for all $N$ tasks
\For{t=1,\ldots,T}
\State Choose task $a_t$ based on $|Q|$ using $\epsilon$-greedy or Boltzmann policy
\State Train \textsc{Student} using task $a_t$ and observe score $o_t = x_t^{(a_t)}$
\State Store score $o_t$ in $D(a_t)$ and timestep $t$ in $E(a_t)$
\State Use linear regression to predict $D(a_t)$ from $E(a_t)$ and use the coef. as $r_t$
\State Update expected return $Q(a_t) = \alpha r_t + (1 - \alpha) Q(a_t)$
\EndFor
\end{algorithmic}
\end{algorithm}

\begin{algorithm}
\caption{Sampling algorithm}\label{sampling_simple}
\begin{algorithmic}
\State Initialize \textsc{Student} learning algorithm
\State Initialize FIFO buffers $D(a)$ with length $K$ for all $N$ tasks
\For{t=1,\ldots,T}
\State Sample reward $\tilde{r}_a$ from $D(a)$ for each task (if $|D(a)|=0$ then $\tilde{r}_a=1$)
\State Choose task $a_t = \argmax_a |\tilde{r}_a|$
\State Train \textsc{Student} using task $a_t$ and observe reward $r_t = x_t^{(a_t)} - x_{t'}^{(a_t)}$
\State Store reward $r_t$ in $D(a_t)$
\EndFor
\end{algorithmic}
\end{algorithm}

\newpage
\section{Batch versions of the algorithms}
\label{appendix:batch_algs}

\begin{algorithm}
\caption{Online algorithm}\label{online_batch}
\begin{algorithmic}
\State Initialize \textsc{Student} learning algorithm
\State Initialize expected return $Q(a)=0$ for all $N$ tasks
\For{t=1,\ldots,T}
\State Create prob. dist. $\vec{a_t}=(p_t^{(1)}, ..., p_t^{(N)})$ based on $|Q|$ using $\epsilon$-greedy or Boltzmann policy
\State Train \textsc{Student} using prob. dist. $\vec{a_t}$ and observe scores $\vec{o_t} = (x_t^{(1)}, ..., x_t^{(N)})$
\State Calculate score changes $\vec{r_t} = \vec{o_t} - \vec{o_{t-1}}$
\State Update expected return $\vec{Q} = \alpha \vec{r_t} + (1 - \alpha) \vec{Q}$
\EndFor
\end{algorithmic}
\end{algorithm}

\begin{algorithm}
\caption{Naive algorithm}\label{online_naive}
\begin{algorithmic}
\State Initialize \textsc{Student} learning algorithm
\State Initialize expected return $Q(a)=0$ for all $N$ tasks
\For{t=1,\ldots,T}
\State Create prob. dist. $\vec{a_t}=(p_t^{(1)}, ..., p_t^{(N)})$ based on $|Q|$ using $\epsilon$-greedy or Boltzmann policy
\State Reset $D(a)=\emptyset$ for all tasks
\For{k=1,\ldots,K}
\State Train \textsc{Student} using prob. dist. $\vec{a_t}$ and observe scores $\vec{o_t} = (x_t^{(1)}, ..., x_t^{(N)})$
\State Store score $o_t^{(a)}$ in list $D(a)$ for each task $a$
\EndFor
\State Apply linear regression to each $D(a)$ and extract the coefficients as vector $\vec{r_t}$
\State Update expected return $\vec{Q} = \alpha \vec{r_t} + (1 - \alpha) \vec{Q}$
\EndFor
\end{algorithmic}
\end{algorithm}

\begin{algorithm}
\caption{Window algorithm}\label{online_window}
\begin{algorithmic}
\State Initialize \textsc{Student} learning algorithm
\State Initialize FIFO buffers $D(a)$ with length $K$ for all $N$ tasks
\State Initialize expected return $Q(a)=0$ for all $N$ tasks
\For{t=1,\ldots,T}
\State Create prob. dist. $\vec{a_t}=(p_t^{(1)}, ..., p_t^{(N)})$ based on $|Q|$ using $\epsilon$-greedy or Boltzmann policy
\State Train \textsc{Student} using prob. dist. $\vec{a_t}$ and observe scores $\vec{o_t} = (x_t^{(1)}, ..., x_t^{(N)})$
\State Store score $o_t^{(a)}$ in $D(a)$ for all tasks $a$
\State Apply linear regression to each $D(a)$ and extract the coefficients as vector $\vec{r_t}$
\State Update expected return $\vec{Q} = \alpha \vec{r_t} + (1 - \alpha) \vec{Q}$
\EndFor
\end{algorithmic}
\end{algorithm}

\begin{algorithm}
\caption{Sampling algorithm}\label{online_sampling}
\begin{algorithmic}
\State Initialize \textsc{Student} learning algorithm
\State Initialize FIFO buffers $D(a)$ with length $K$ for all $N$ tasks
\For{t=1,\ldots,T}
\State Sample reward $\tilde{r}_a$ from $D(a)$ for each task (if $|D(a)|=0$ then $\tilde{r}_a=1$)
\State Create one-hot prob. dist. $\vec{\tilde{a}_t}=(p_t^{(1)}, ..., p_t^{(N)})$ based on $\argmax\nolimits_a |\tilde{r}_a|$
\State Mix in uniform dist. : $\vec{a_t} = (1 - \epsilon) \vec{\tilde{a}_t} + \epsilon/N$
\State Train \textsc{Student} using prob. dist. $\vec{a_t}$ and observe scores $\vec{o_t} = (x_t^{(1)}, ..., x_t^{(N)})$
\State Calculate score changes $\vec{r_t} = \vec{o_t} - \vec{o_{t-1}}$
\State Store reward $r_t^{(a)}$ in $D(a)$ for each task $a$
\EndFor
\end{algorithmic}
\end{algorithm}

\clearpage
\section{Decimal Number Addition Training Details}
\label{appendix:addition}

Our reimplementation of decimal addition is based on Keras \citep{chollet2015keras}. The encoder and decoder are both LSTMs with 128 units. In contrast to the original implementation, the hidden state is not passed from encoder to decoder, instead the last output of the encoder is provided to all inputs of the decoder. One curriculum training step consists of training on 40,960 samples. Validation set consists of 4,096 samples and 4,096 is also the batch size. Adam optimizer \citep{kingma2014adam} is used for training with default learning rate of 0.001. Both input and output are padded to a fixed size.

In the experiments we used the number of steps until 99\% validation set accuracy is reached as a comparison metric. The exploration coefficient $\epsilon$ was fixed to 0.1, the temperature $\tau$ was fixed to 0.0004, the learning rate $\alpha$ was 0.1, and the window size $K$ was 10 in all experiments.
 
\section{Minecraft Training Details}
\label{appendix:minecraft}

The Minecraft task consisted of navigating through randomly generated mazes. The maze ends with a target block and the agent gets 1,000 points by touching it. Each move costs -0.1 and dying in lava or getting a timeout yields -1,000 points. Timeout is 30 seconds (1,500 steps) in the first task and 45 seconds (2,250 steps) in the subsequent tasks.

For learning we used the \textit{proximal policy optimization} (PPO) algorithm \citep{schulman2017proximal} implemented using Keras \citep{chollet2015keras} and optimized for real-time environments. The policy network used four convolutional layers and one LSTM layer. Input to the network was $40\times 30$ color image and outputs were two Gaussian actions: move forward/backward and turn left/right. In addition the policy network had state value output, which was used as the baseline. Figure \ref{f14} shows the network architecture.

\begin{figure}[h]
  \includegraphics[scale=0.4]{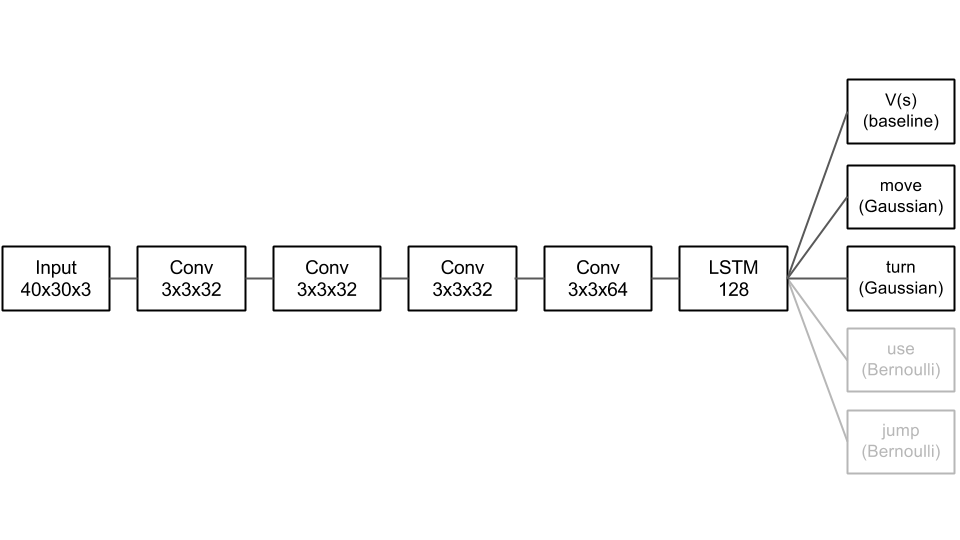}
\caption{Network architecture used for Minecraft.}
\label{f14}
\end{figure}

For training we used a setup with 10 parallel Minecraft instances. The agent code was separated into runners, that interact with the environment, and a trainer, that performs batch training on GPU, similar to \cite{babaeizadeh2016reinforcement}. Runners regularly update their snapshot of the current policy weights, but they only perform prediction (forward pass), never training. After a fixed number of steps they use FIFO buffers to send collected states, actions and rewards to the trainer. Trainer collects those experiences from all runners, assembles them into batches and performs training. FIFO buffers shield the runners and the trainer from occasional hiccups. This also means that the trainer is not completely on-policy, but this problem is handled by the importance sampling in PPO.

\begin{figure}[h]
  \includegraphics[scale=0.4]{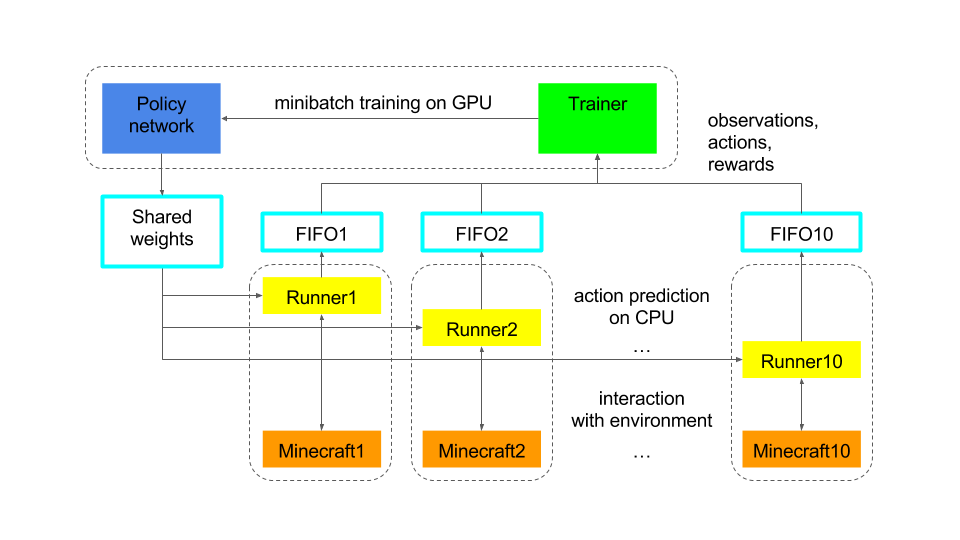}
\caption{Training scheme used for Minecraft.}
\label{f14}
\end{figure}

During training we also used frame skipping, i.e. processed only every 5th frame. This sped up the learning considerably and the resulting policy also worked without frame skip. Also, we used auxiliary loss for predicting the depth as suggested in \citep{mirowski2016learning}. Surprisingly this resulted only in minor improvements.

For automatic curriculum learning we only implemented the Window algorithm for the Minecraft task, because other algorithms rely on score change, which is not straightforward to calculate for parallel training scheme. Window size was defined in timesteps and fixed to 10,000 in the experiments, exploration rate was set to 0.1.

The idea of the first task in the curriculum was to make the agent associate the target with a reward. In practice this task proved to be too simple - the agent could achieve almost the same reward by doing backwards circles in the room. For this reason we added penalty for moving backwards to the policy loss function. This fixed the problem in most cases, but we occasionally still had to discard some unsuccessful runs. Results only reflect the successful runs.

We also had some preliminary success combining continuous (Gaussian) actions with binary (Bernoulli) actions for "jump" and "use" controls, as shown on figure \ref{f14}. This allowed the agent to learn to cope also with rooms that involve doors, switches or jumping obstacles, see \url{https://youtu.be/e1oKiPlAv74}.

\end{appendices}

\end{document}